\documentclass[letterpaper, 10 pt, conference]{ieeeconf}  

\usepackage{framed,multirow}
\usepackage{booktabs}
\usepackage{amsmath} 
\usepackage{amssymb}  
\usepackage{graphicx}
\usepackage{url}
\usepackage{balance}

\title{\LARGE \bf
GazeTarget360: Towards Gaze Target Estimation in 360-Degree for Robot Perception
}

\author{Zhuangzhuang Dai$^{1*}$, Vincent Gbouna Zakka$^{2}$, Luis J. Manso$^{3}$, and Chen Li$^{4}$\\
\normalsize $^{1,2,3}$Dept. of Applied AI and Robotics,
        Aston University, Birmingham, United Kingdom\\
\normalsize $^{4}$Dept. of Materials and Production, Aalborg University, DK-9220, Aalborg East, Denmark\\
{$^{*}$Corresponding address: \tt\small z.dai1@aston.ac.uk}
}

\begin{document}

\maketitle
\thispagestyle{empty}
\pagestyle{empty}

\begin{abstract}

Enabling robots to understand human gaze target is a crucial step to allow capabilities in downstream tasks, for example, attention estimation and movement anticipation in real-world human-robot interactions. Prior works have addressed the in-frame target localization problem with data-driven approaches by carefully removing out-of-frame samples. Vision-based gaze estimation methods, such as OpenFace, do not effectively absorb background information in images and cannot predict gaze target in situations where subjects look away from the camera. In this work, we propose a system to address the problem of 360-degree gaze target estimation from an image in generalized visual scenes. The system, named GazeTarget360, integrates conditional inference engines of an eye-contact detector, a pre-trained vision encoder, and a multi-scale-fusion decoder. Cross validation results show that GazeTarget360 can produce accurate and reliable gaze target predictions in unseen scenarios. This makes a first-of-its-kind system to predict gaze targets from realistic camera footage which is highly efficient and deployable. Our source code is made publicly available at: \url{https://github.com/zdai257/DisengageNet}.

\end{abstract}

\section{INTRODUCTION}

\begin{figure}[tpb]
      \centering
      \includegraphics[width=0.49\textwidth]{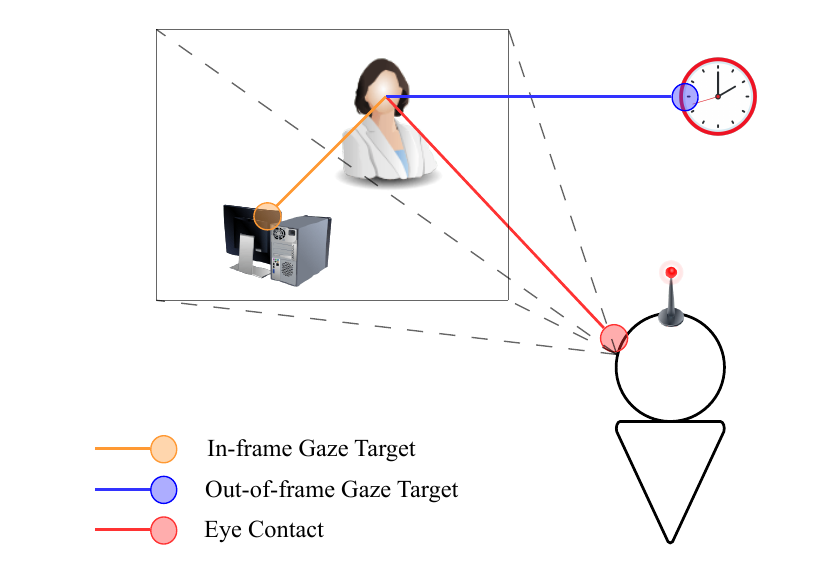}
      \caption{Classification of region of gaze target. Research have addressed tasks in each region separately but jointly. Our GazeTarget360 system can predict all three gaze target regions and reliably estimate in-frame gaze target location.}
      \label{gazetarget}
   \end{figure}

Estimating human attention is paramount for robots in real-world interactions. Gaze contains crucial information about humans' intentions and potential actions. There have been a stream of research investigating human gaze direction, eye contact, and attended targets through vision-based techniques. In human-robot interaction (HRI), detection and tracing of human users have been prevalently integrated in modern robots~\cite{gazecontrol2020}. However, it is yet challenging to utilize such systems in robots to effectively predict human attention in real-world settings. Humans may gaze at out-of-frame targets beyond robots' immediate field-of-view, as shown in Fig.~\ref{gazetarget}. Furthermore, `eye contact' between humans and a robotic agent is a crucial indicator of interest~\cite{dai_icac2023}, engagement~\cite{engageintensity2019}, and intent to interact~\cite{gazedriverintent2019} which are under-explored in state-of-the-art robotics research.

Existing vision-based gaze estimation methods separately study eye contact (EC), attention target, and facial landmarks which are too fragmented to be useful in robotics. Eye contact detector~\cite{onfocus2021} seeks to infer whether a user is gazing at the robotic agent. Joint attention (JA)~\cite{attendedtarget2020} posits two or more people intentionally sharing focus on a common object or activity. Eye landmarks and gaze direction estimation~\cite{openface2016} can represent human affects but remain agnostic of the scene context or attended targets. These research problems are always being looked at individually. Robots are expected to possess all above capabilities of estimating gaze targets to thrive in real-world interactions.

To this end, our goal is to enable reliable and unrestricted gazed target estimation in a unified framework as shown in Fig.~\ref{gazetarget}. Specifically, human gaze may land at in-frame targets (IFT), or out-of-frame targets (OFT) beyond a visible sensor's immediate field-of-view~\cite{attendedtarget2020}. Having mutual eye contact (EC) with a robot is a very special case of OFT, which should be separately detected for robots~\cite{deepeyecontact2020}. We propose a generalized system, GazeTarget360 (GT360), to identify all three scenarios. In GT360, we leverage several state-of-the-art large pre-trained Deep Learning models, including a vision foundation model~\cite{dinov2} incorporating all-purpose features for zero-shot downstream tasks and an EC convolutional model supervised by millions of face samples. A commonly used human face detector~\cite{dlib2009} is integrated as a frontend sensor to activate the conditional inference of EC detection and IFT/OFT estimation. We propose a novel multi-scale fusion module to learn fine-grained and global features for enhanced eye gaze and target representation learning. Combining the foundation model, the fusion module, and a compact learnable decoder, our GT360 system realizes competitive performance on a range of unseen datasets.

Our major contributions include (1) a first-of-its-kind system to freely predict gaze targets in 360-degree for real-world robot perception; (2) an enhanced encoder-decoder model with a multi-scale fusion module for robot IFT/OFT prediction with outstanding efficiency to train; (3) unifying existing gaze target related datasets and annotations for strategic training and comprehensive evaluation. Specifically, our GT360 system is the first work to synthesize the gaze target output space putting EC, OFT, and IFT under a unified framework. We show that EC is a special kind of gaze target bearing significance for real-world scene understanding and GT360 can reliably detect them. Our multi-scale fusion module represents a crucial contribution to efficient cross-scale attention facilitating feature fusion at different spatial granularities.

GT360 allows gaze target estimation from a 2D image input in arbitrary camera angles, which is important for real-world applications. This is particularly valuable in realistic applications where our system can be plugged-in and play. Our system has demonstrated state-of-the-art performance on EYEDIAP~\cite{eyediap2014} and zero-shot EC tasks, and competitiveness upon GazeFollow~\cite{gazefollow2015} and VideoAttentionTarget~\cite{attendedtarget2020} with a unified and more challenging output space. By addressing the problem of human gaze targets beyond robots' immediate field of view, this research has the potential to significantly advance robot perception system design and human-robot interaction (HRI) applications. The source code is made publicly available at: \url{https://github.com/zdai257/DisengageNet}.

\section{RELATED WORK}

\subsection{Machine Vision for Human Gaze}

Automated attention estimation has become increasingly important in various applications, including smart education~\cite{DAiSEE2016}, human-robot interaction (HRI)~\cite{dai_icac2023}, and Advanced Driving Assistance Systems (ADAS)~\cite{gazedriverintent2019}. Gaze tracking has emerged as a fundamental component of attention estimation. Gaze information provides crucial insights into human agents' attentional focus and cognitive states. Machine Vision based approach~\cite{gazefollow3d2023,goo2021} to gaze estimation has seen remarkable development in offering contactless and automatic gaze tracking solutions in the past decade.

Recasens~\cite{gazefollow2015} formulated the research problem of gaze following and published the first large-scale dataset of images. Chong~\cite{attendedtarget2020} extended the merits to videos and, importantly, incorporated binary classification of in and out-of-frame gaze targets. Ryan~\cite{gazelle2024} proposed a neat encoder-decoder model for gaze target estimation achieving state-of-the-art performance on all aforementioned benchmarks. Detecting whether a subject has eye-contact with the camera has realized robust performance~\cite{deepeyecontact2018,onfocus2021}. Another domain of research focuses on gaze direction estimation from facial appearance features. For instance, gaze rays can be deduced from cropped face or eye regions~\cite{openface2016,gaze360_2019}.

Despite many endeavours, the research problems of gaze following, eye contact detection, and attended target localization are usually studied separately. This makes understanding human attention in real-world settings a significant challenge. The gap lies in bridging the eye feature engineering and scene understanding in a unified system. Traditionally, accurate eye contact detection usually requires a high resolution in the cropped eye region~\cite{columbia2013,eyediap2014}. This is difficult for low-cost visible sensors or when subjects are far. Inspired by the stream of research in gaze following, we leverage large pre-trained vision foundation models~\cite{dinov2} to incorporate head, gesture, as well as visual saliency in the scene to enable generalized gaze target estimation. Different from Gaze360~\cite{gaze360_2019} which predicts vectorized gaze directions but ignoring context, our approach is target-oriented encompassing the scene context to directly predict attended targets.

\subsection{Deep Models for Gaze Estimation}

Recasens~\cite{gazefollow2015} used dual convolutional encoders for a full-image pathway and a head pathway. Chong~\cite{attendedtarget2020} utilized ConvLSTM module and cross attention to solve the gaze following in continuous video streams. Recently, Gaze-LLE~\cite{gazelle2024} demonstrates that head encoding pathway is unnecessary and realizes state-of-the-art performance with just a pre-trained foundation visual encoder. This shows large pre-trained visual encoders have learned head, gesture, eye, and saliency features quite well if being decoded appropriately.

Since gaze estimation requires detecting extremely subtle cues (such as slight head movements or eye positions), low-level features from earlier layers may capture these fine details more effectively than the more abstract, high-level features. Gaze target is also correlated to a subject's head pose, gesture, and global saliency in images~\cite{dai_egocap2023}. Herein, we use a large pre-trained DINOv2~\cite{dinov2} encoder in our system for a generalized high- and low-level feature representation. We incorporate features from various fields and fuse them to obtain a multi-scale representation that provides a richer context for the subsequent layers of the model. We also integrate an enhanced ViT~\cite{vit2020} decoder architecture to bridge the gap of accurate gaze target localization.



\section{METHOD}

\begin{figure*}[tp]
    \centerline{\includegraphics[width=0.96\textwidth]{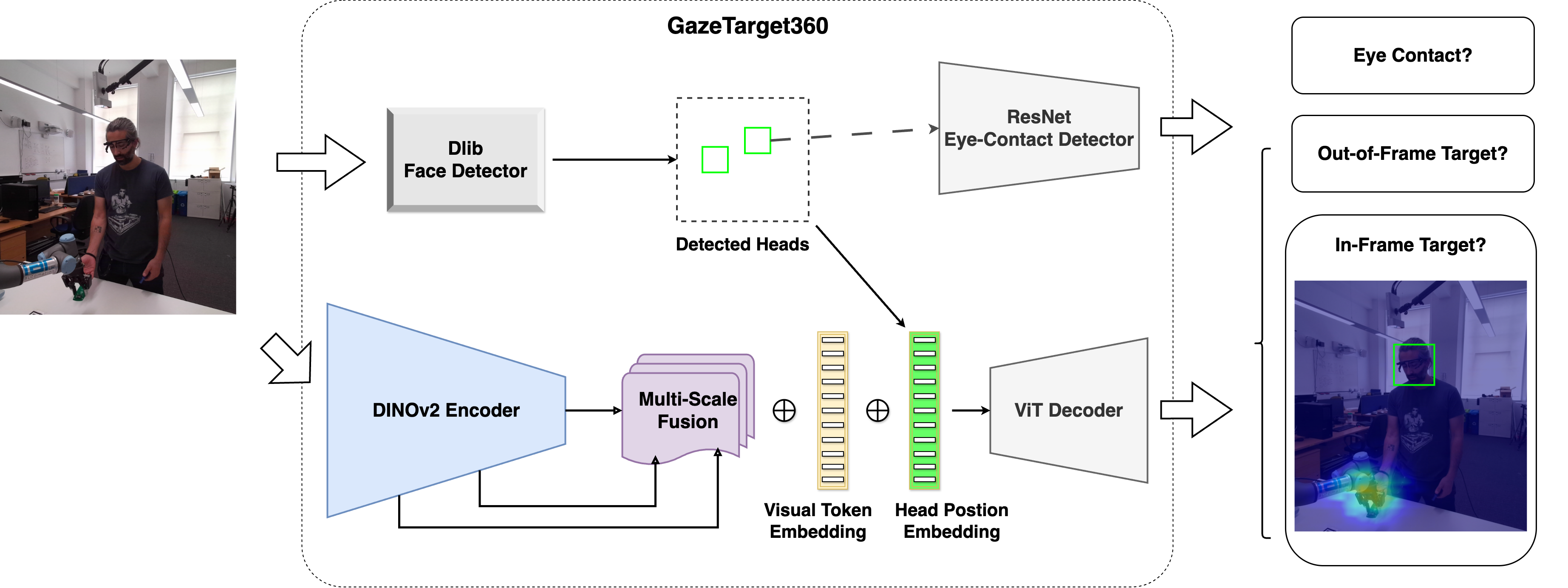}}
    \caption{Overall architecture of our proposed GazeTarget360 system. The detected heads will inflict eye contact detection. If non-eye-contacting is decided, the gaze target estimation engine will process the image consuming full background as contextual information. A multi-scale fusion (MSF) module utilizes multi-scale tokenization to aggregate three-stage receptive fields for fine-grained gaze and target features. This makes a first-of-its-kind 360-degree gaze target estimator.}
    \label{fig:system}
\end{figure*}

We propose GazeTarget360 (GT360) for unrestricted gaze target estimation from any visible data. A state-of-the-art face detector from dlib~\cite{dlib2009} is used as sensor to trigger the rest of the pipeline, and provides the faces detected that are used as bounding box prompts. The OFT/IFT prediction engine will subsequently process each non-eye-contacting head prompt to locate a gaze target. The system architecture is outlined in Fig.~\ref{fig:system}.

\subsection{Problem Formulation}

The gaze target system is expected to freely estimate any subject's gaze target from a colour image, $I \in \mathbb{R}^{h\times w\times 3}$. Specifically, the system should discern the case where a subject's impinging gaze vector intersects the camera pose; it should localize the target position in pixel coordinates if a subject is gazing at IFT; and it should tell the cases of OFT. The formula can be expressed as,

\begin{equation}
  output(I) =
    \begin{cases}
      1 & \text{if}\,\, P_{EC} \ge \sigma_{}\\
      0 & \text{if}\,\, P_{EC} < \sigma_{} \,\text{and}\,\, P_{IFT} < 0.5\\
      M & \text{if}\,\, P_{EC} < \sigma_{} \,\text{and}\,\, P_{IFT} \ge 0.5
    \end{cases}       
\end{equation}

\noindent
where $P_{EC}$ and $P_{IFT}$ are the probabilities of eye-contact (EC) and gazing at in-frame target (IFT), respectively; $\sigma$ is the cut-off probability to determine EC and we find $0.85$ a robust threshold through experiments. The output is of two-stage: a classification head, $y$, determining EC, non-EC with an OFT, or non-EC with an IFT represented as a heatmap, $M$, containing gaze target probabilities within the frame $I$.

\subsection{Eye Contact Classification}

An RGB image may contain multiple people. We leverage a commonly used \textit{dlib} face detector~\cite{dlib2023} to extract all heads, $[x_0, x_1, ..., x_n] = f(I)$, where $n$ is the number of heads. The cropped head regions will act as both the source for EC detection and the prompts for OFT/IFT estimation. In the first stage of the conditional inference, an estimate of EC probability ranged between $(0, 1)$ will be produced, which can be expressed as;

\begin{equation}
    y = H(I, x_k)
\end{equation}

Eye contact with the camera is a special case of gaze target. Wherein, this rare gaze direction relative to a robot's visible sensor contains crucial cues of interest and intent of engagement. Vision-based EC detection has been well-explored by the community. It is generally agreed by prior work that an EC case is independent from the background~\cite{columbia2013}, meaning an EC can be accurately detected by cropped pixels of eyes~\cite{deepeyecontact2018}. E. Chong~\cite{deepeyecontact2020} developed a robust EC binary classifier through supervised learning with 4 million annotated faces. Based on this prior work, we construct an EC classification module by taking as input the cropped heads. We use the pre-trained ResNet as backbone with parameters learned in $H(\cdot)$.

\subsection{Gaze Target Localization}

In the second stage, we classify OFT or IFT leveraging an encoder-decoder architecture. We combine a ViT~\cite{vit2020} decoder with multi-scale fusion which extracts fine-grained gaze and target features with a large pre-trained visual encoder. The head bounding boxes, $x_k$, from previous stage are used as head position prompts. This stage can be formulated as

\begin{equation}
    y, M = G(I, f(I))
\end{equation}

\noindent
where $G(\cdot)$ is the encoder-decoder model that jointly classifies \textit{in} or \textit{out} gaze region and predicts target positional probabilities in a heatmap if the former holds true.

Inspired by Ryan~\cite{gazelle2024}, the decoder comprises of a head prompt channel with 2D positional encoding of the head positions, token embeddings, ViT blocks, and two-head outputs. We adopt two fully-connected layers to for the IFT/OFT classification head and stacked convolutional layers for the heatmap head. We use DINOv2~\cite{dinov2} as the scene encoder as has proven optimal performance in the previous work. 

We introduce a multi-scale fusion module to effectively integrate information from different spatial scales. Instead of using a single fixed token size, multi-scale tokenization extracts three different patch sizes, of scaling factors $1$, $0.5$, and $0.25$, to create embeddings at different spatial granularities (Fig.~\ref{fig:system}). A convolutional layer with $1\times1$ kernel size is used to align the output channel dimensions for fusion. This allows the model to capture both fine-grained details of human eyes and global context of salient targets. A single ViT block is used to construct a lightweight decoder which shows competitive performance compared to state-of-the-art.

In OFT cases, the model will inflict a zero masking to the  heatmap head to suppress backpropagation. If a gaze target is outside the field of view, a gaze direction vector may be generated, e.g., using OpenFace~\cite{openface2016}, to register direction of pursuit for further motion planning. In the case of IFT, the output is a heatmap of $M_{64\times 64}$ grids each containing probabilities of gaze target. This resolution aligns with prior work for the ease of evaluation. With an IFT, the heatmap will directly highlight the region of interest containing a subject's attended target. This will facilitate a range of robotic applications such as grasping target prediction, joint-attention evaluation, and future behaviour anticipation.

\subsection{Training}

A key challenge of unrestricted gaze target estimation is missing annotated data of all target locations displayed in Fig.~\ref{gazetarget}. We bridge this gap by merging datasets each covering a subspace of the target labels. The GazeFollow~\cite{gazefollow2015} dataset is an early work of IFT localization with large-scale data. The VideoAttentionTarget~\cite{attendedtarget2020} dataset provides IFT coordinates from diverse video sources as well as binary OFT/IFT labels.

We pre-train the IFT pathway with GazeFollow before fine-tuning on VideoAttentionTarget. We first train the model on GazeFollow with 15 epochs following the author's recommended settings. Then, we fine-tune it on VideoAttentionTarget with a 5-epoch warm-up stage followed by 10 epochs training with a \textit{lr} of $1e-5$ and a cosine \textit{lr} decay. We use AdamW optimizer, a batch size of 32, and data augmentation techniques including colour jitter, random gray-scaling, and uniformly resizing input to $(448, 448)$. Note that during training the DINOv2 encoder parameters remain frozen. Our model has 1.94M learnable parameters which is significantly more efficient than 2.93M in Gaze-LLE~\cite{gazelle2024}.

We use pixel-wise binary cross-entropy loss for the pre-training, and an additional binary cross-entropy loss for the fine-tuning stage which can be written as,

\begin{equation}
    \mathcal{L} = \mathcal{L}_{(64, 64)} + \lambda \cdot \mathcal{L}_{BCE}
\end{equation}

\noindent
where $\lambda$ is a real scalar balancing the target localization and binary classification tasks. We find $\lambda=1.0$ yields best performance for GT360. We apply Gaussian blurs to the ground-truth heatmap labels to soften the target loss.

To sum up, we propose GT360 system combining a powerful pre-trained eye-contact conditional inference engine and a frozen DINOv2~\cite{dinov2} frontend to encode global scene features as well as fine-grained head and eye features. We develop a multi-scale fusion module in GT360 to enhance efficiency and information fusion at different spatial scales.

\begin{table}[!b]
\centering
\caption[stats]{Datasets used for training and/or evaluation.}
\label{table:datasets}
\begin{tabular}{c|ccccc} 
    \multirow{2}{*}{\textbf{Dataset}} & \multicolumn{2}{c}{\textbf{No. of Samples}} & \multicolumn{3}{c}{\textbf{Annotation}} \\
     & \textit{Train} & \textit{Test} & EC & OFT & IFT\\
    \hline
    GazeFollow~\textsuperscript{\textdagger}~\cite{gazefollow2015} & 117K & 4,782 &  &  & \checkmark  \\
    VideoAttentionTarget~\cite{attendedtarget2020} & 58,507 & 13,127 &  & \checkmark & \checkmark  \\
    ColumbiaGaze~\cite{columbia2013} & \multicolumn{2}{c}{5,880} & \checkmark & \checkmark &   \\
    MPIIFaceGaze~\cite{mpiigaze2019} & \multicolumn{2}{c}{37,667} & \checkmark & \checkmark &  \\
    EYEDIAP~\cite{eyediap2014} & \multicolumn{2}{c}{1,750 snapshots} &  & \checkmark & \checkmark  \\
    WALI-HRI~\cite{dai_walihri2024} & \multicolumn{2}{c}{5h video} &  & \checkmark &   \\
    \hline
\end{tabular}
\vspace*{2.5mm}

\noindent{\textdagger Downloadable file from official site was corrupted. Data accessed through HuggingFace \url{https://huggingface.co/datasets}. Sample size may vary.}
\end{table}

\begin{figure*}[!t]
    \centerline{\includegraphics[width=0.99\textwidth]{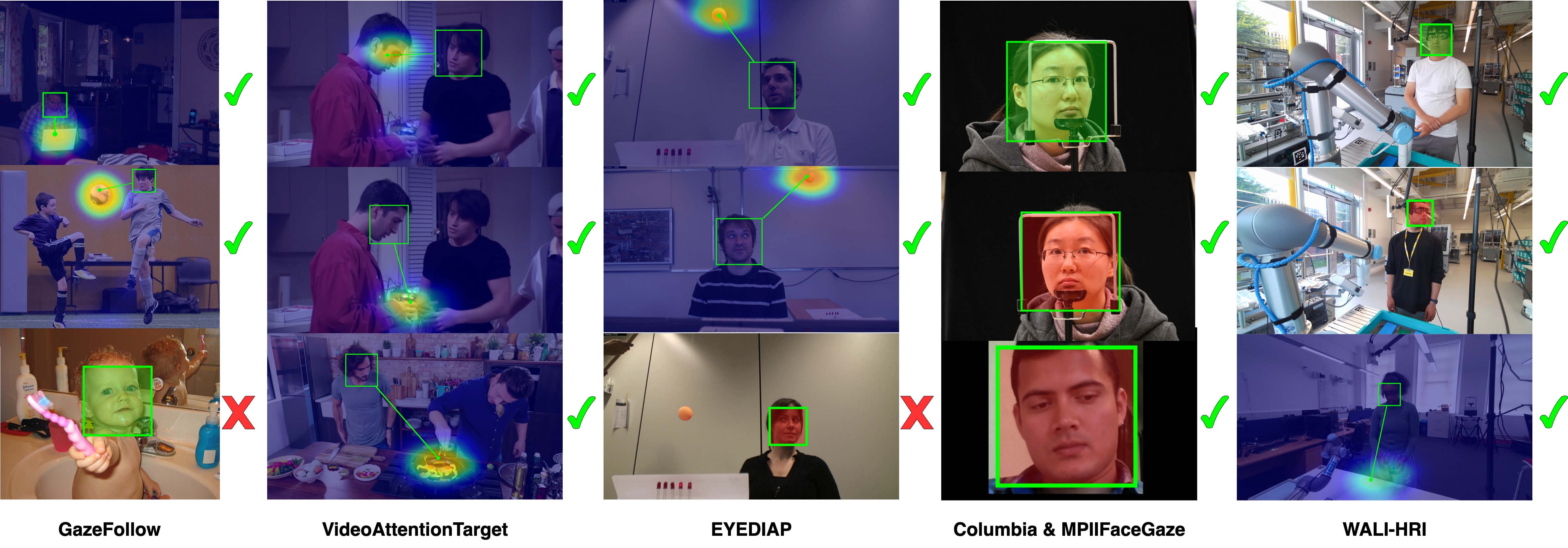}}
    \caption{Qualitative evaluation results. Green boxes indicate detected heads. A green rendering represents eye contact (EC); a red rendering represents gazing at out-of-frame target (OFT); an arrow pointing toward a green dot represents in-frame target (IFT) location with an overlaying heatmap. The GT360 system produces consistent performance across multiple unseen datasets. A GazeFollow sample at the bottom row conflicts with the label which is debatable. GT360 sometimes fails at extreme eye angles such as in the EYEDIAP sample.}
    \label{fig:demo}
\end{figure*}

\begin{table*}[!t]
\centering
\caption[methods]{Evaluation results on GazeFollow and VideoAttentionTarget.}
\label{table:methods}
\begin{tabular}{lcccccc} \toprule
    & \multirow{2}{*}{No. of Learnable Parameters} & \multicolumn{2}{c}{GazeFollow} & \multicolumn{3}{c}{VideoAttentionTarget} \\
    Method &  & AUC$\uparrow$ & Mean L2$\downarrow$ & AP$_{in/out}\uparrow$ &  AUC$\uparrow$ & Mean L2$\downarrow$ \\ \midrule
    Recasens et al.~\cite{gazefollow2015} & 50M & 0.878 & 0.19 & - & - & - \\
    Chong et al.~\cite{attendedtarget2020} & 61M & 0.921 & 0.137 & 0.853 & 0.86 & 0.134 \\
    Tafasca et al.~\cite{tafasca2024} & 135M & 0.944 & 0.113 & 0.891 & - & 0.107 \\
    Gaze-LLE$_{base}$~\cite{gazelle2024} & 2.8M & 0.956 & 0.104 & 0.897 & 0.933 & 0.107 \\
    Gaze-LLE$_{large}$~\cite{gazelle2024} & 2.9M & 0.958 & 0.099 & 0.903 & 0.937 & 0.103 \\ \midrule
    GT360 & 1.94M & 0.957 & 0.101 & 0.887 & 0.934 & 0.103 \\
    \bottomrule
\end{tabular}
\end{table*}

\section{EXPERIMENTS}

Existing datasets, as shown in Table~\ref{table:datasets}, provide partial labels of all three kinds: eye contact (EC), out-of-frame target (OFT), and in-frame target positions (IFT). We exploit GazeFollow~\cite{gazefollow2015} and VideoAttentionTarget~\cite{attendedtarget2020} for training our OFT/IFT module following the method in~\cite{gazelle2024} whilst reserving test splits for benchmarking. We use the rest datasets and their available labels for out-of-distribution evaluation. We report the system performance on EYEDIAP~\cite{eyediap2014} for IFT precision, ColumbiaGaze~\cite{columbia2013} and MPIIFaceGaze~\cite{mpiigaze2019} for EC robustness, and WALI-HRI~\cite{dai_walihri2024} for qualitative evaluation in real-world HRI scenarios.

We evaluate our system using the following metrics: average precision (AP) of OFT/IFT classification, precision, recall, and F1 score of EC/non-EC classification, area under the curve (AUC) of IFT heatmap probabilities, and mean error distance (mean L2) of IFT pixels.

\subsection{Comparison to state-of-the-art methods}

\noindent
\textbf{In-frame target precision.} We evaluate the GT360 OFT/IFT module on the test splits of GazeFollow and VideoAttentionTarget, as shown in Table~\ref{table:methods}. In comparison to state-of-the-art methods, our proposed system demonstrates competitive performance. The GT360 model has the least number of learnable parameters making it the least computationally expensive to train.

The EYEDIAP~\cite{eyediap2014} dataset contains videos of subjects continuously gazing at a floating target which went both out-of-frame and in-frame with ground-truth projected on-screen position. We evenly sample 50 frames from all floating target videos and construct a dataset of 1,750 samples including $38.6\%$ OFT cases where the floating target was moved beyond field of view. It can be seen from Table~\ref{table:eyediap}, GT360 outperforms the state-of-the-art method with or without accurate head prompts. We notice Gaze-LLE does not seem to benefit from a larger encoder-decoder architecture. This is probably because the backgrounds of EYEDIAP samples are rather plain. Our multi-scale fusion module can attend to crucial gaze cues disregarding uninteresting features to remain efficacious in this out-of-distribution challenge.

\begin{table}[t]
\centering
\caption[methods]{Evaluation results on EYEDIAP dataset.}
\label{table:eyediap}
\begin{tabular}{lccc} \toprule
    & \multicolumn{3}{c}{EYEDIAP}  \\
    Method & AP$_{in/out}\uparrow$ & AUC$\uparrow$ & Mean L2$\downarrow$  \\ \midrule
    Gaze-LLE$_{b}$ & 0.725 & 0.617 & 0.411 \\
    Gaze-LLE$_{l}$ & 0.614 & 0.596 & 0.421  \\
    Gaze-LLE$_{b}$ + head prompt & 0.73 & 0.597 & 0.423  \\
    Gaze-LLE$_{l}$ + head prompt & 0.662 & 0.593 & 0.431  \\ \midrule
    GT360 & 0.756 & 0.593 & 0.314  \\
    \bottomrule
\end{tabular}
\end{table}

\noindent
\textbf{Eye-contact robustness.} To assess the eye-contact module~\cite{deepeyecontact2020}, we process the 3D gaze data offered in ColumbiaGaze~\cite{columbia2013} and MPIIFaceGaze~\cite{mpiigaze2019} datasets to label eye-contact cases. In \cite{mpiigaze2019}, the 3D positions of a subject's face centre ($\mathbf{fc}$), and ground-truth gaze target positions ($\mathbf{gt}$), are provided. A 3D gaze vector can be derived $\mathbf{v}=\mathbf{gt}-\mathbf{fc}$ with its unit vector $\mathbf{d}$. We compute the distance between the camera origin and the gaze vector by subtracting the gaze vector with its projection on the unit direction, $\text{dist}=\left\| \mathbf{v} - \left( \mathbf{v} \cdot \mathbf{d} \right) \mathbf{d} \right\|$. If \text{dist} is smaller than 30mm, we label the sample as a true EC. The ColumbiaGaze~\cite{columbia2013} dataset offers ground-truth head angles of 56 subjects. We take the samples of $0^{\circ}$ elevation and $0^{\circ}$ yaw angles as true ECs.

The EC precision, recall, and F1 scores are reported compared to baseline methods as shown in Table~\ref{table:ec}. We compare our GT360 EC module to a Generative Adversarial Network inspired method, SSLEC~\cite{deepeyecontact2018}, and a deep convolution model, DEEPEC~\cite{deepec2017}. The comparative methods are trained on ColumbiaGaze and MPIIFaceGaze, respectively. Then, the models are evaluated on the unseen subjects' samples with a leave-one-out scheme. The results show that GT360 can produce accurate and robust EC predictions making it a reliable frontend detector in the conditional inference framework.

\begin{table}[t]
  \caption{Evaluation results on EC / non-EC classification.}
  \label{table:ec}
  \centering
  \begin{tabular}{@{}llll@{}}
    \toprule
     & \multicolumn{3}{c}{ColumbiaGaze} \\
    {Method} & {Precision} & {Recall} & {F1-score} \\
    \midrule
    SSLEC & 0.7993 & 0.8242 & 0.7921 \\
    DEEPEC & 0.8846 & 0.8783 & 0.8859 \\
    GT360 & 0.9091 & 0.9416 & 0.925  \\
    \toprule
     & \multicolumn{3}{c}{MPIIFaceGaze} \\
    \midrule
    DEEPEC & 0.5503 & 0.129 & 0.1962  \\
    GT360 & 0.6857 & 0.8086 & 0.6634 \\
  \bottomrule
  \end{tabular}
\end{table}

\subsection{Qualitative evaluation}

We qualitatively evaluate GT360 on all aforementioned datasets including WALI-HRI~\cite{dai_walihri2024} which provides 5h recordings of 26 subjects engaging in a human-robot co-assembly task. As shown in Fig.~\ref{fig:demo}, our system shows excellent real-world adaptability across four unseen datasets. Accurate classification of EC/OFT/IFT and precise localization of IFT are demonstrated. Note that the bottom GazeFollow sample of a baby is classified as EC by GT360 although the original annotation indicates an IFT of toothbrush. It is debatable if the baby was engaged by the cameraperson or truly gazing at the toothbrush. This poses new questions to review the existing annotations to align the gaze target output space with our framework. An EYEDIAP sample is wrongly classified as OFT. The EYEDIAP datasets contains many footage of extreme gaze angles making it a challenge for current approach.

\section{LIMITATIONS}

This research addresses the problem of gaze target estimation for realistic human-robot interactions, including joint detections of IFT/OFT/EC. We evaluate GT360 on diverse datasets and demonstrate its robust performance in unrestricted gaze target space. However, GT360 is not a true 3D gaze target solver since its 3D spatial reasoning is confined by what representations a frozen DINOv2 encoder has learned. Per-class performance analysis is not yet possible due to a lack of labelled data explicitly separating IFT/OFT/EC. Inheriting the same backbone of Gaze-LLE, GT360 has the ability to infer gaze target solely based on gesture and context with only side or back of a subject's head visible. Nonetheless, the performance will be limited by visibility factors like occlusion and poor illumination, which curse any vision-only systems. In future work, we will investigate gaze target depth, as well as informed motion planning of robotic head and perception sensors.






\section*{ACKNOWLEDGMENT}

Experiments were run on Aston Engineering and Physical Science (EPS) Machine Learning Server, funded by the EPSRC Core Equipment Fund, Grant EP/V036106/1.


\bibliographystyle{IEEEtrans}
\bibliography{ref}

\end{document}